\documentclass[runningheads]{llncs}

\usepackage{eccv}

\usepackage{eccvabbrv}
\usepackage{graphicx}
\usepackage{booktabs}
\usepackage{amsmath}
\usepackage{xcolor}
\usepackage[accsupp]{axessibility}

\usepackage{hyperref}

\title{Geometry Beats Estimated Depth: RGB-Only Multi-Camera 3D Tracking under Sim2Real}

\titlerunning{Geometry Beats Estimated Depth for Multi-Camera 3D Tracking}

\author{Abdullah Naeem\inst{1} \and Ayon Dey\inst{1}\textsuperscript{\$} \and Anav Katwal\inst{1}\textsuperscript{\$} \and
Md Tamjidul Hoque\inst{1}\textsuperscript{*} \and Noman Khan\inst{2}}

\authorrunning{A. Naeem et al.}

\institute{LSU New Orleans, New Orleans, USA\\
\email{\{anaeem,adey,akatwal,thoque\}@lsuneworleans.edu}
\and
PinPark, Inc.\\
\email{noman@pinpark.co}}

\begin{document}
\maketitle
{\let\thefootnote\relax\footnotetext{\textsuperscript{\$}Equal contribution.\quad\textsuperscript{*}Corresponding author.}}

\begin{abstract}
The AI City Challenge 2026 Track 1 evaluates multi-camera 3D perception in large indoor warehouses under a synthetic-to-real (Sim2Real) setting; depth is available only for training and validation, so inference is RGB-only. We use two RGB-only routes as a controlled test of one hypothesis: that \emph{cross-view geometric consistency}, not monocular depth accuracy, governs performance under Sim2Real. The first is a \emph{geometry-first} pipeline: YOLO11x detection, homography lifting to the world frame, class-level 3D size priors, multi-camera fusion, world-coordinate tracking, and offline tracklet stitching. The second is \emph{estimated-depth pseudo-LiDAR}: monocular depth (D4RT, Metric3D~v2) back-projected into a fused point cloud and passed to a 3D detector (V-DETR), mirroring prior point-cloud winners that used \emph{provided} depth. The gap is decisive: geometry-first reaches 13.0 3D HOTA (51.6 LocA), whereas pseudo-LiDAR collapses to 0.12 (9.2 LocA). We trace the collapse to cross-view inconsistency of monocular depth---scale correction is necessary but not sufficient---which domain-adaptation fine-tuning does not repair within budget. Within the geometry pipeline, offline stitching is the only intervention that helps; SAHI detection, appearance Re-ID, learned lifting, RT-DETR ensembling, test-time augmentation, and domain randomization all fail to beat the baseline detector. The bottlenecks are complementary: detection quality bounds the geometry route (DetA), localization consistency bounds pseudo-LiDAR (LocA). We release a complete, reproducible RGB-only pipeline and ablation.

\end{abstract}

\section{Introduction}

Multi-camera 3D perception in industrial indoor environments is a challenging setting for detection, localization, and identity association. AI City Challenge 2026 Track 1 requires participants to detect and track people and mobile objects, including forklifts, mobile robots, humanoids, transporters, and pallet trucks, across synchronized warehouse cameras. The output is a single text file containing world-coordinate 3D bounding boxes and object identities for each frame. The challenge is especially difficult because the training data is primarily synthetic while hidden test scenes include real-world videos and visual stressors. In addition, depth maps are available only for training and validation; inference must rely on RGB images.

Our goal was to build a complete end-to-end system quickly, evaluate the importance of each component, and identify the main bottleneck. We implemented a practical baseline around a strong 2D detector and explicit camera geometry. The system detects objects in each camera view, lifts each detection into a world coordinate system using calibration metadata and homographies, assigns class-prior 3D dimensions, fuses duplicate detections across cameras, and performs world-coordinate tracking. This pipeline produced a valid leaderboard submission and enabled systematic ablation.

The main conclusion from our experiments is that the current system is detection-limited. Our best submission achieved a stable localization score near 51 LocA and an association score higher than detection accuracy, but DetA remained low. Attempts to improve identity association or replace the detector did not improve final HOTA. This suggests that sparse 2D detection followed by heuristic lifting is not sufficient for high-ranking performance in this challenge. We therefore tested the natural alternative---estimated-depth pseudo-LiDAR, the route used by prior provided-depth winners---and found it performs far \emph{worse}, isolating cross-view depth consistency rather than detection alone as the obstacle for learned-geometry substitutes.

\paragraph{Central hypothesis.} We frame the two routes as a controlled test of a single question: for RGB-only multi-camera 3D tracking under Sim2Real conditions, is \emph{cross-view geometric consistency}---agreement of the recovered geometry across cameras---more decisive than the \emph{per-image accuracy} of monocular depth? The geometry-first lift maximizes cross-view consistency by construction (all cameras share one calibrated ground plane) while discarding fine per-pixel depth; estimated-depth pseudo-LiDAR maximizes per-image depth detail but recovers each camera's geometry independently, sacrificing cross-view consistency. Because both routes address the same task, data, and metric, comparing them isolates which factor governs performance. \emph{Our hypothesis is that cross-view geometric consistency, not monocular depth accuracy, is the dominant factor}---and the experiments below support it: the consistency-preserving lift reaches 13.0 HOTA while the depth-accurate but cross-view-inconsistent pseudo-LiDAR route collapses to 0.12.

Our contributions are primarily \emph{scientific findings}, supported by a reproducible system:

\begin{itemize}
    \item \textbf{A systematic geometry-vs-depth comparison.} A controlled head-to-head of geometry-first lifting against estimated-depth pseudo-LiDAR for RGB-only Sim2Real 3D tracking, showing that explicit cross-view geometry outperforms learned monocular depth by roughly two orders of magnitude (13.0 vs 0.12 HOTA), and identifying cross-view metric consistency---not per-image depth accuracy and not the 3D detector---as the governing factor.
    \item \textbf{A geometric-consistency diagnostic.} A simple, annotation-free \emph{floor-coherence} metric that quantifies cross-view geometric consistency, separates scale error from consistency error, and predicts the pseudo-LiDAR collapse before any detector is run.
    \item \textbf{A design principle from ablation.} We group every intervention we tried into detector-side, geometry-side, and association-side categories and give the common reason each category fails to move HOTA: the geometry route is bounded by detection \emph{quality}, so only association-side relinking (offline stitching) helps, while adding detections (detector-side) or replacing calibrated geometry with learned substitutes (geometry-side) hurts.
    \item \textbf{A reproducible baseline.} A complete, open RGB-only multi-camera 3D tracking pipeline for AI City Challenge 2026 Track 1 and its full ablation, as a reference point for Sim2Real 3D perception.
\end{itemize}

\section{Related Work}

\paragraph{Multi-camera 3D tracking.} The AI City Challenge series has driven progress in multi-target multi-camera tracking and, more recently, scene-level 3D perception~\cite{naphade2023aicity,aicity2026track1}. Strong prior entries fuse per-camera observations into a common 3D frame and associate identities over time, typically via tracking-by-detection with motion and appearance cues~\cite{wojke2017deepsort,zhang2022bytetrack}; recent point-cloud winners voxelize fused depth into a scene representation for 3D detection and offline tracklet linking. Our setting differs in that depth is unavailable at inference, so geometry must be recovered from RGB and calibration alone.

\paragraph{Monocular depth and pseudo-LiDAR.} Pseudo-LiDAR~\cite{wang2019pseudolidar} back-projects estimated depth into a point cloud so that LiDAR-style 3D detectors can operate on images. Monocular depth estimation has advanced rapidly~\cite{ranftl2020midas,yang2024depthanything}: recent models predict metric depth zero-shot (Metric3D~\cite{yin2023metric3d}) or reconstruct dynamic scenes from video (D4RT~\cite{zhang2025d4rt}). Once back-projected, such geometry is consumed by point-cloud 3D detectors, whether voting-based~\cite{qi2019votenet} or transformer-based~\cite{misra2021_3detr}. We test whether such \emph{estimated} depth can replace the \emph{provided} depth used by earlier winners, and find that cross-view inconsistency---not raw per-image depth accuracy---is the limiting factor (Section~\ref{sec:pseudolidar}).

\paragraph{2D detection and association.} We build on the YOLO family~\cite{redmon2016yolo,ultralytics2026}, compare against transformer detectors (DETR~\cite{carion2020detr}, RT-DETR~\cite{zhao2024rtdetr}), and evaluate sliced inference (SAHI)~\cite{akyon2022sahi} for small-object recall. Association uses geometry-based world-coordinate tracking with an offline tracklet-stitching stage, and we evaluate with 3D HOTA~\cite{luiten2021hota}, which factors detection, association, and localization accuracy.

\section{Task and Evaluation}

Each submission is a plain-text file with one row per predicted 3D detection:

\begin{equation}
(\text{scene}, \text{class}, \text{id}, \text{frame}, x, y, z, w, l, h, \theta).
\end{equation}

The challenge evaluates seven classes: Person, Forklift, NovaCarter, Transporter, FourierGR1T2, AgilityDigit, and PalletTruck. The official metric is 3D HOTA~\cite{luiten2021hota}, which balances detection accuracy, association accuracy, and localization accuracy. We report the leaderboard components HOTA, DetA, AssA, and LocA.

The challenge setting is Sim2Real. Synthetic training data is available with 2D/3D annotations, calibration, maps, and depth maps. The hidden test set contains additional real or stressed scenes. Since depth is not available at inference time, our system uses only RGB videos and calibration for test prediction.

\section{Method}

\begin{figure}
\centering
\includegraphics[width=\textwidth]{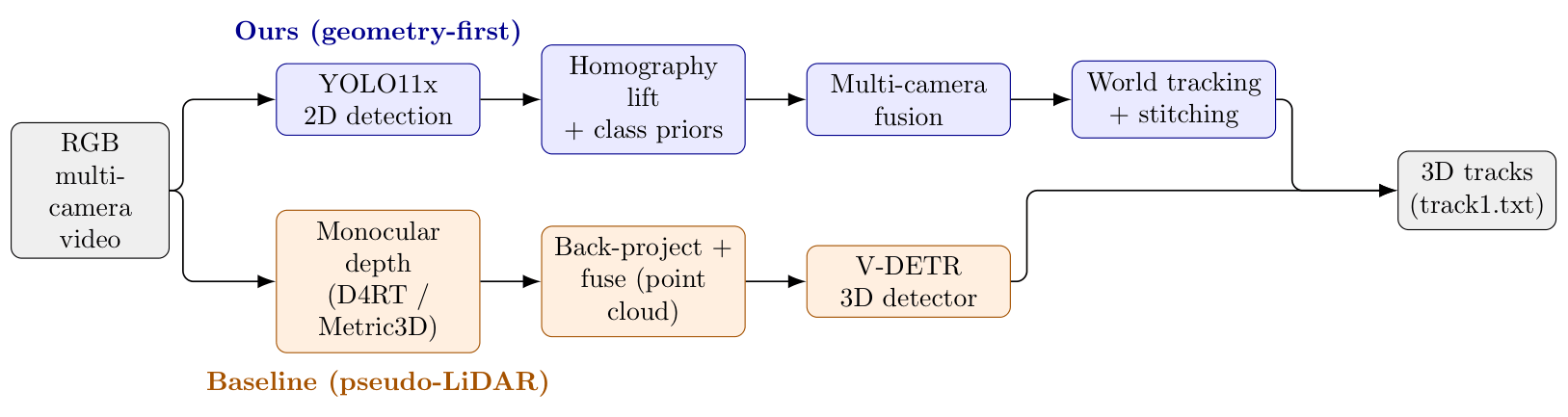}
\caption{The two RGB-only routes we compare. \emph{Top (ours):} per-camera 2D detection, homography ground-plane lifting with class-size priors, same-frame multi-camera fusion, world-coordinate tracking, and offline tracklet stitching. \emph{Bottom (baseline):} monocular depth, back-projection and multi-camera fusion into a scene point cloud, and a V-DETR 3D detector.}
\label{fig:pipeline}
\end{figure}

We compare two RGB-only routes (Fig.~\ref{fig:pipeline}): a geometry-first pipeline (Sections~\ref{subsec:det}--\ref{subsec:stitch}) and a pseudo-LiDAR baseline (Section~\ref{subsec:plb}).

\subsection{2D Detection}
\label{subsec:det}

We trained an Ultralytics YOLO11x detector at 1280 input resolution. The model predicts per-camera 2D bounding boxes and class labels. Detections are cached as JSONL files per scene and camera. We use a low detector confidence threshold during inference to retain recall, then apply stronger filtering in the tracker and submission post-processing.

\subsection{World-Frame Lifting}

Each 2D detection is lifted into the world coordinate system using the camera calibration and a homography-based footpoint projection. For a bounding box, the bottom-center point is projected into the ground/world plane. We found the following projection setting to be best on validation diagnostics:

\begin{quote}
\texttt{--invert-homography --footpoint-y 1.00 --world-plane xy}
\end{quote}

This configuration produced a median validation lift error of approximately 0.633 m in our alignment diagnostic. The projected point defines the object position in the world plane. The vertical coordinate and 3D box dimensions are assigned using class-level priors estimated from training and validation annotations. Yaw is estimated heuristically.

\subsection{Multi-Camera Fusion}

Since the same object can be visible in multiple synchronized camera views, detections are first fused per frame and class in world space. Detections within a class-specific spatial radius are merged into one scene-level observation. This reduces duplicate boxes before tracking and partially recovers detections from cameras with weaker views.

\subsection{World-Coordinate Tracking}

\begin{figure}[t]
\centering
\includegraphics[width=\linewidth]{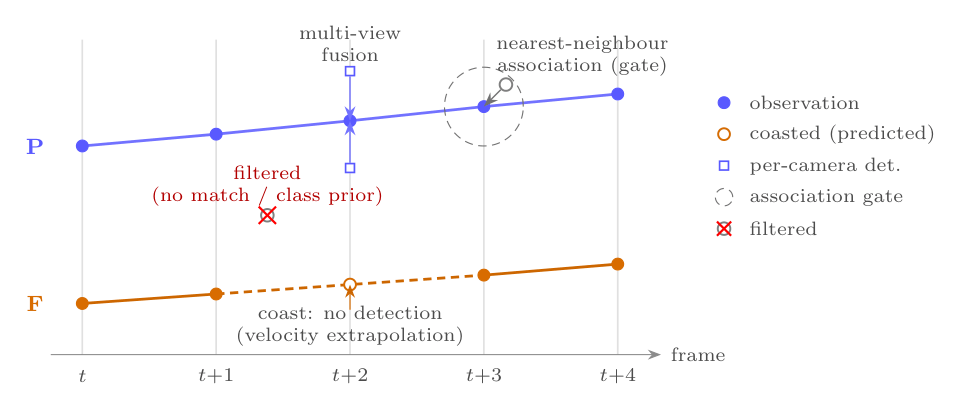}
\caption{World-frame association and track maintenance. Per-camera detections are lifted to the ground plane and fused across views (squares$\,\to\,$dot); each fused observation is linked to the nearest active track whose predicted position lies within a distance gate (dashed circle). When a track has no detection in a frame it is \emph{coasted} by extrapolating its velocity (open marker, dashed segment) for up to a fixed number of frames before termination; observations that match no track or violate class priors are filtered ($\textcolor{red}{\times}$). Person (\textbf{P}) and Forklift (\textbf{F}) tracks are shown over five frames.}
\label{fig:tracker}
\end{figure}

Tracking is performed separately for each class (Fig.~\ref{fig:tracker}). The tracker associates fused detections to active tracks using world-coordinate distance and simple motion prediction: each fused observation is matched to the nearest active track whose predicted position falls within a distance gate. We include optional coasting to bridge short gaps---when a track has no detection in a frame, its position is extrapolated from velocity for up to a fixed number of frames before termination. After tracking, short tracks and observations that match no track or violate class priors are filtered out, and object IDs are remapped to satisfy the submission constraints.

\subsection{Offline Tracklet Stitching}
\label{subsec:stitch}

The online tracker fragments identities whenever an object is missed for longer than the coasting window and later reappears, which caps association accuracy. We add an offline stitching stage that links same-class fragments: fragment $B$ is joined to fragment $A$ when $B$ begins shortly after $A$ ends (within a gap budget) and $B$'s start position lies within a distance threshold of $A$'s velocity-extrapolated endpoint. Links are matched one-to-one (cheapest residual first) and chained via union-find, and each merged chain receives a single scene-unique identity. Because stitching only relabels identities and never alters boxes, it improves association without changing detection or localization: in our experiments it raises AssA from 14.53 to 16.71 at essentially unchanged DetA and LocA. This stage makes the submission offline, since it uses future track evidence.

\subsection{Pseudo-LiDAR Baseline}
\label{subsec:plb}

For comparison we implement the estimated-depth pseudo-LiDAR route. Per-camera depth is predicted from RGB with a monocular model---D4RT~\cite{zhang2025d4rt} or Metric3D~v2~\cite{yin2023metric3d}---back-projected using calibration, and fused into a world-frame point cloud. A V-DETR 3D detector~\cite{shen2024vdetr} trained on the provided-depth clouds then predicts 3D boxes; we additionally fine-tune it on estimated-depth clouds to reduce the train/test depth-domain gap. Section~\ref{sec:pseudolidar} reports the outcome.

\section{Experiments}

\subsection{Detector Performance}

Table~\ref{tab:yolo-val} shows YOLO11x validation performance. The detector has high precision but limited recall, especially for PalletTruck. This recall weakness directly limits DetA and final HOTA.

\begin{table}
\centering
\caption{YOLO11x validation detector metrics.}
\label{tab:yolo-val}
\begin{tabular}{lrrrr}
\toprule
Class & Precision & Recall & mAP50 & mAP50-95 \\
\midrule
All & 0.907 & 0.603 & 0.655 & 0.486 \\
Person & 0.817 & 0.723 & 0.753 & 0.577 \\
Forklift & 0.812 & 0.601 & 0.646 & 0.438 \\
NovaCarter & 0.989 & 0.806 & 0.840 & 0.709 \\
Transporter & 0.968 & 0.647 & 0.729 & 0.521 \\
PalletTruck & 0.950 & 0.239 & 0.309 & 0.185 \\
\bottomrule
\end{tabular}
\end{table}

On the training split, class-wise detector metrics were much higher, with recall above 0.87 for all evaluated classes and above 0.98 for PalletTruck. This indicates that the model can learn the object categories, but generalization to validation/test appearance is weak. We attribute this to domain shift, small or distant objects, real camera artifacts, illumination changes, and low object-background contrast.

\subsection{Leaderboard Results and Ablations}

Table~\ref{tab:leaderboard-ablation} summarizes the main official leaderboard experiments. The YOLO11x geometry baseline with offline tracklet stitching is the best result: stitching raises AssA from 14.53 to 16.71 (and HOTA from 12.49 to 13.04) at unchanged DetA/LocA, since it only relinks fragmented identities. Re-ID, learned lifting, detector ensembling, test-time augmentation, and YOLO26/domain-randomized training all reduced final performance.

\begin{table}
\centering
\caption{Official leaderboard experiments.}
\label{tab:leaderboard-ablation}
\begin{tabular}{lrrrrl}
\toprule
Experiment & HOTA & DetA & AssA & LocA & Decision \\
\midrule
YOLO11x + geometry baseline & 12.4891 & 10.7900 & 14.5291 & 51.5784 & Base \\
\quad + offline tracklet stitching & 13.0413 & 10.7897 & 16.7105 & 51.5785 & \textbf{Best} \\
Light Re-ID variant & 10.5120 & 10.5182 & 13.0722 & 51.0577 & Reject \\
Low-confidence recall bump & 12.0249 & 10.2352 & 14.2158 & 51.5209 & Reject \\
Learned lift MLP & 0.9418 & 0.9679 & 0.8150 & 39.0014 & Reject \\
YOLO + RT-DETR ensemble & 10.8819 & 9.6225 & 12.5183 & 50.6654 & Reject \\
YOLO11x TTA & 11.1725 & 9.9193 & 12.2902 & 51.4127 & Reject \\
SAHI sliced detection (+ stitching) & 11.4943 & 7.2698 & 17.2804 & 49.8864 & Reject \\
YOLO26 + stress augmentation & $\sim$10 & -- & -- & -- & Reject \\
\bottomrule
\end{tabular}
\end{table}

Rather than list the interventions individually, we group them by the pipeline stage they target---detector, geometry, or association. The outcomes cluster cleanly by category, and each category fails (or succeeds) for a single, generalizable reason.

\paragraph{Detector-side interventions (all reject).} RT-DETR gave small validation recall gains for Forklift and PalletTruck but was worse overall; a YOLO+RT-DETR ensemble added noisy boxes and lowered HOTA; YOLO26 trained on a domain-randomized~\cite{tobin2017domainrand} stress-augmented set also underperformed the YOLO11x baseline; test-time augmentation and a low-confidence recall bump likewise did not help. Sliced inference (SAHI), which tiles each frame to recover small and distant objects, more than doubled the detection count but \emph{lowered} DetA ($10.79\!\to\!7.27$) and HOTA ($13.04\!\to\!11.49$). \emph{Common reason:} under Sim2Real the deficit is detection \emph{quality}, not quantity or detector family---adding boxes (SAHI, low threshold, ensembling) trades precision for recall and floods false positives, while swapping architectures does not transfer the missing real-domain generalization.

\paragraph{Geometry-side interventions (all reject).} Replacing the calibrated homography lift with a learned bbox-to-3D MLP failed severely (HOTA $0.94$, LocA $39.0$), and the estimated-depth pseudo-LiDAR route of Section~\ref{sec:pseudolidar} collapsed entirely (HOTA $0.12$). \emph{Common reason:} both replace the one component that is already reliable---explicit, cross-view-consistent geometry from shared calibration---with a learned substitute that lacks metric grounding and cross-view agreement. A crop/bbox-only regressor carries too little information for metric 3D, and monocular depth is cross-view inconsistent (Section~\ref{sec:pseudolidar}).

\paragraph{Association-side interventions (one helps).} Appearance Re-ID~\cite{ye2021reid} from object crops did not improve association and reduced HOTA, because real/stress scenes yield low-resolution, self-similar, domain-shifted crops that make appearance embeddings less reliable than geometry. The one intervention that helped is offline tracklet stitching (Section~\ref{subsec:stitch}), which raises AssA $14.53\!\to\!16.71$ (HOTA $12.49\!\to\!13.04$) at unchanged DetA/LocA. \emph{Common reason:} association is \emph{not} the bottleneck, so adding an appearance signal only injects noise, whereas repairing fragmentation---relabeling identities without touching boxes---recovers the single kind of association error the online tracker actually makes.

\paragraph{Design principle.} Across all three categories, one variable explains the outcomes: the geometry route is limited by detection quality, and its explicit geometry is already its most reliable part. The interventions that help are exactly those that improve association \emph{without} disturbing detection or geometry; those that hurt either add low-quality detections or replace trustworthy geometry with a learned substitute. For RGB-only Sim2Real 3D perception this yields a concrete principle---\emph{invest in detection quality and preserve explicit geometry; do not trade precision for recall, and do not replace calibrated geometry with learned depth or lifting}---which also predicts the pseudo-LiDAR result of Section~\ref{sec:pseudolidar}.

\section{Qualitative Analysis}

\begin{figure}
\centering
\includegraphics[width=0.85\textwidth]{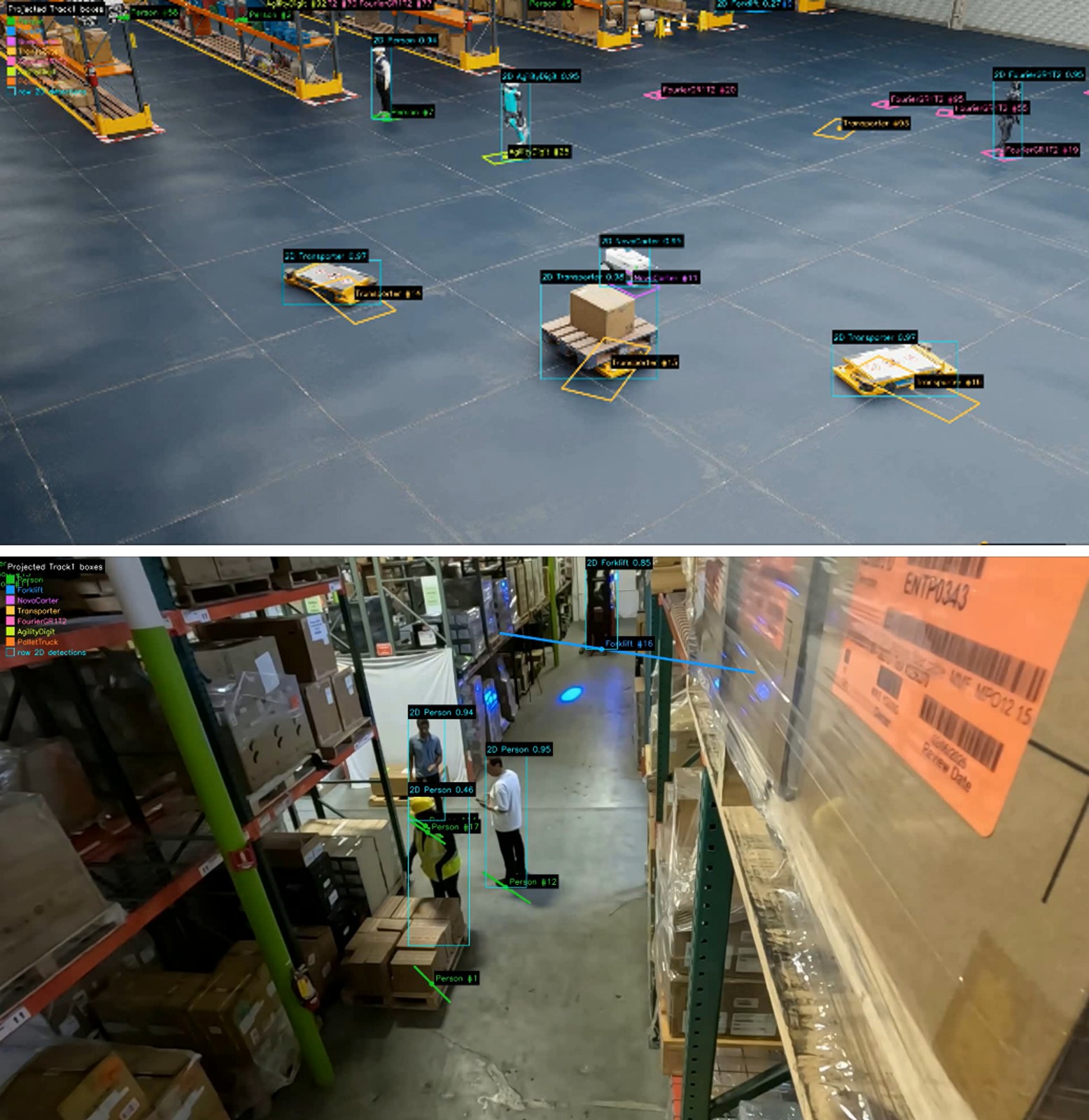}
\caption{World-frame track boxes (projected back to the image) overlaid with the raw 2D detections, on a synthetic scene (top) and a real scene (bottom). \emph{Top:} the projected boxes over-detect (false positives) and their ground-plane bases are mis-placed relative to the objects---homography-lift error. \emph{Bottom:} the 2D detector correctly finds people, but these detections fail to propagate into world-frame tracks, so the world output \emph{under-covers} real objects. The 2D$\to$3D lift is the weak link in complementary ways---spurious, mis-placed boxes on synthetic and dropped detections on real---both bounding DetA. Best viewed in colour.}
\label{fig:overlay}
\end{figure}

We built video overlays that combine world-frame track boxes with the raw 2D detections (Fig.~\ref{fig:overlay}); they expose the 2D-to-3D lift as the weak link, and in domain-dependent ways. On the synthetic scene the world boxes \emph{over-detect}, and their ground-plane bases are frequently mis-placed relative to the objects (lift error), producing false positives. On the real scene the failure inverts: the 2D detector still finds people, but those detections often fail to propagate into world tracks, so the world output \emph{under-covers} real objects. In both cases the error lies in producing clean world-frame detections, not in association---where a correct world detection exists, the geometry tracker maintains a usable trajectory.

\begin{figure}[t]
\centering
\includegraphics[width=0.60\textwidth]{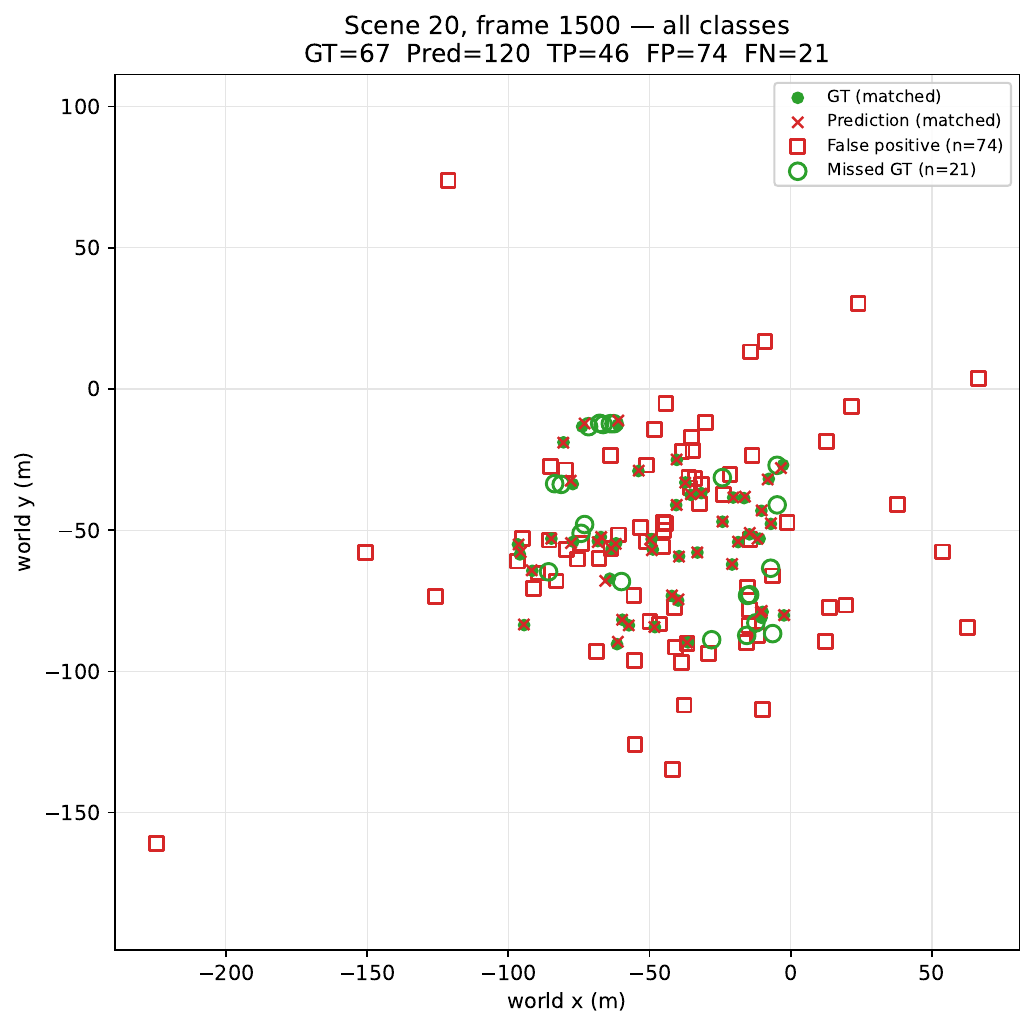}
\caption{World-coordinate predictions (red) vs.\ ground truth (green) for the final submission on a validation frame (Scene~20, frame~1500, all classes). Linked pairs are matches within 2\,m; matched pairs align reasonably (moderate localization; LocA is the strongest of the three components), but predictions far outnumber ground truth ($\mathrm{Pred}=120$ vs $\mathrm{GT}=67$), with false positives (74) exceeding true positives (46)---the over-detection that bounds DetA. A few predictions are gross lift/coasting errors outside the scene (off-view). Best viewed in colour.}
\label{fig:bev}
\end{figure}

The score decomposition supports this qualitative finding. Our LocA is relatively stable near 51, and AssA is higher than DetA. Thus, the main limitation is not identity association alone but the availability and quality of scene-level detections. Figure~\ref{fig:bev} makes this concrete on a validation frame: matched predictions align reasonably with ground truth (LocA is the comparatively strongest component), yet predictions far outnumber ground truth ($\mathrm{Pred}=120$ vs $\mathrm{GT}=67$; 74 false positives against 46 true positives), the over-detection that bounds DetA.

\paragraph{Synthetic-to-real gap.} The test set mixes synthetic scenes (Warehouses~23--25) and real-world scenes (Warehouses~26--27). Detection recall is markedly lower on the real scenes, consistent with the large train-vs-validation detector gap, so the real scenes contribute disproportionately to missed detections and to the DetA ceiling. This localizes the dominant error source to real-domain detection rather than to lifting or association.\footnote{All scores reported in this paper are those returned by the evaluation server's public leaderboard, which ranks submissions on approximately half of the test set; final scores on the full test set may differ. Per-scene HOTA and recall breakdowns are deferred to the camera-ready version.}

\paragraph{Where detection errors originate.} Three axes localize the detection bottleneck using evidence already in the paper. \emph{By class:} validation recall ranges from $0.81$ (NovaCarter) down to $0.24$ (PalletTruck) (Table~\ref{tab:yolo-val}); the low-profile, self-similar PalletTruck and the partially-occluded Forklift ($0.60$) dominate false negatives, whereas compact, high-contrast robots are detected reliably. \emph{By error type:} on the validation frame of Fig.~\ref{fig:bev} false positives outnumber true positives ($74$ vs $46$) while $46$ ground-truth objects are simultaneously missed---the detector both over-fires (mis-lifted or duplicated boxes) and under-covers (small, distant, or occluded instances), both symptoms of detection \emph{quality} rather than association. \emph{By domain:} recall is markedly lower on real than on synthetic scenes, so real-domain instances contribute disproportionately to missed detections. The three views converge: DetA is bounded by detection quality for small, thin, and real-domain objects, not by lifting or association. A finer breakdown by absolute object distance and occlusion level would sharpen this further and is limited only by the availability of per-object distance/occlusion annotations.

\paragraph{Robustness and variability.} Inference in our pipeline is deterministic, so run-to-run seed variance is not the relevant axis; the meaningful variability is \emph{across classes and domains}, and it is already substantial. Per-class detector recall is $0.60\pm0.20$ across the five evaluated classes (Table~\ref{tab:yolo-val})---from $0.81$ (NovaCarter) down to $0.24$ (PalletTruck)---so a single aggregate score hides which categories drive the DetA ceiling. Performance also splits sharply by domain, with markedly lower recall on real than on synthetic scenes. Per-scene 3D~HOTA breakdowns across the validation warehouses are deferred to the camera-ready version.

\section{Estimated-Depth Pseudo-LiDAR: A Negative Result}
\label{sec:pseudolidar}

The natural RGB-only route to true 3D perception is \emph{pseudo-LiDAR}: estimate per-pixel depth from RGB, back-project each camera into a common world frame, fuse into a scene point cloud, and run a 3D point-cloud detector---exactly the recipe used by prior point-cloud winners, but with \emph{provided} depth. Because 2026 forbids depth at inference, we tested whether \emph{estimated} depth can substitute. We report the outcome as a controlled negative result, since it is the most-assumed path to a stronger 3D representation.

\paragraph{Setup.} We instantiate the pseudo-LiDAR baseline of Section~\ref{subsec:plb} with D4RT and Metric3D~v2 depth and the provided-depth-trained V-DETR detector, evaluated with and without domain-adaptation fine-tuning on estimated-depth clouds.

\begin{figure}
\centering
\includegraphics[width=0.80\textwidth]{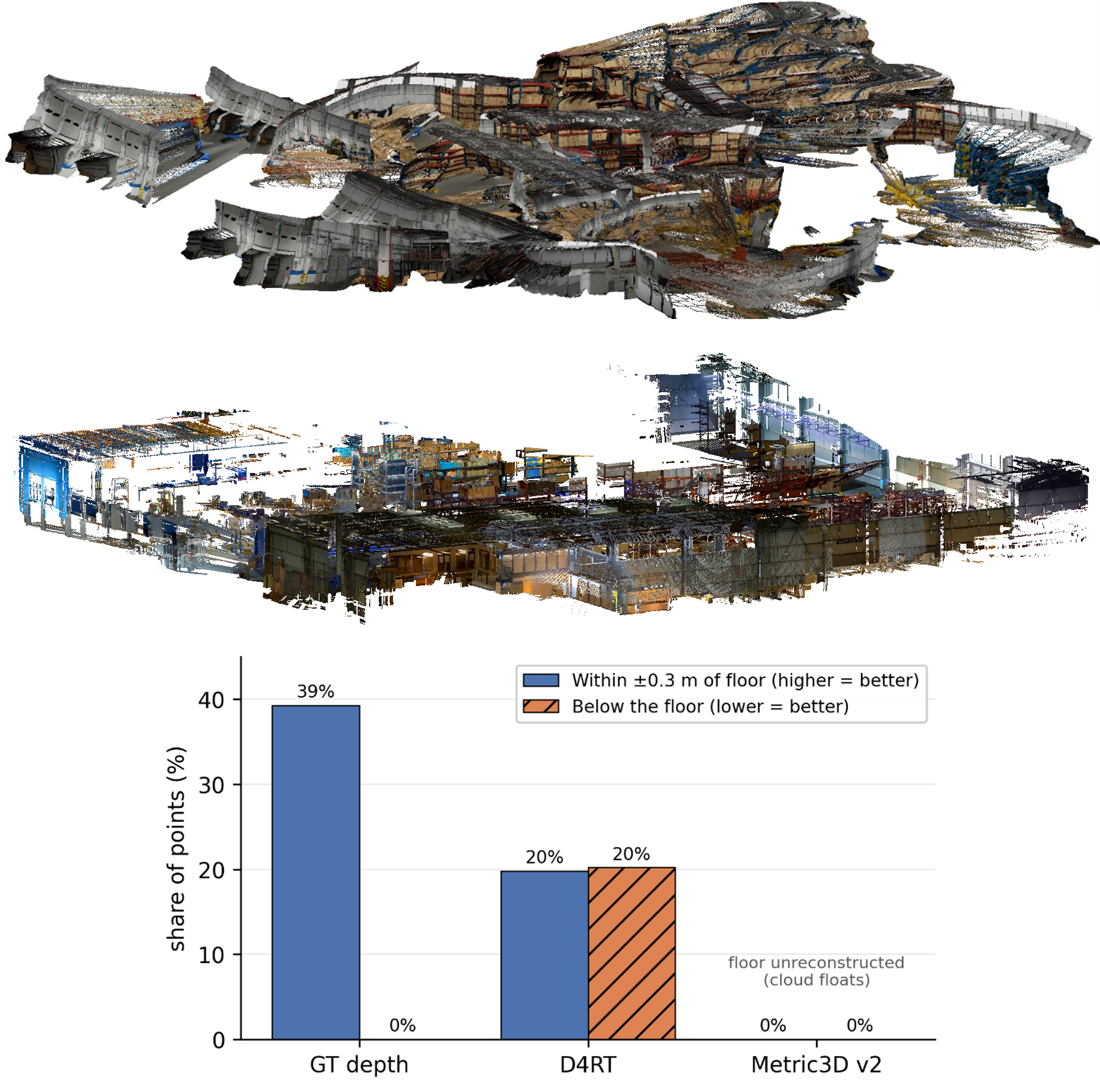}
\caption{Why estimated-depth pseudo-LiDAR fails. \emph{Top:} a fused cloud from scale-corrected D4RT (estimated) depth---warped, fragmented, and non-planar. \emph{Middle:} the provided-depth (ground-truth) cloud for the same scene, with a flat floor and coherent structure. \emph{Bottom:} floor coherence---share of points within $\pm0.3$~m of the ground plane, and share below it---for GT vs.\ D4RT vs.\ Metric3D~v2, quantifying the cross-view inconsistency that collapses LocA.}
\label{fig:depth}
\end{figure}

\paragraph{Findings.} \textbf{(i) Scale.} Monocular depth is not metric out of the box: D4RT required a near-constant $\sim$4.3$\times$ correction to reach metric scale, while Metric3D~v2 is metric by construction but places clouds inconsistently---its reconstructed floor floats above the ground plane. \textbf{(ii) Cross-view inconsistency.} Even after scale correction, monocular depth disagrees across views, so multi-camera fusion yields warped, non-planar geometry (Fig.~\ref{fig:depth}). As a proxy we measure \emph{floor coherence} on a validation scene: provided-depth clouds place 39\% of points within $\pm0.3$~m of the floor with 0\% below it, whereas scale-corrected D4RT places only 20\% in-band with 20\% \emph{below} the floor, and Metric3D leaves the floor essentially unreconstructed. \textbf{(iii) Result.} The provided-depth-trained V-DETR applied to estimated-depth clouds scored 0.12 HOTA with 9.2 LocA---roughly two orders of magnitude below the geometry lift---and domain-adaptation fine-tuning on estimated-depth clouds did not yield usable detections within our compute budget.

\paragraph{Why floor coherence?} The warehouse floor is a large, planar surface at known elevation that is visible to most cameras, which makes it an ideal probe for \emph{global} cross-view consistency: if per-camera depth agrees across views, every camera's floor points must land at the same world height, so their concentration in a thin band around the ground plane---and the absence of points below it---directly measures cross-view metric agreement, with no object annotation required. It is also a \emph{necessary} condition for correct boxes, since every target rests on the floor, so a warped floor guarantees mis-placed objects. We set the band to $\pm0.3$~m to match object scale: it is below the height of the shortest targets yet within the vertical tolerance at which a predicted box still overlaps ground truth, and the qualitative ordering (provided\,$\gg$\,D4RT\,$\gg$\,Metric3D) is insensitive to the exact value over $0.2$--$0.5$~m, since Metric3D reconstructs essentially no floor and the provided cloud has essentially no sub-floor mass regardless of the cutoff. We prefer this proxy to a per-point error against ground-truth depth, which is unavailable at test time and, more importantly, measures \emph{per-image} accuracy rather than cross-view consistency---monocular depth can be locally accurate yet globally warped, which is exactly the failure we observe. Plane-fit residuals or inter-view reprojection error are alternatives, but they require per-camera correspondence; floor coherence needs only the calibrated ground height and is therefore simpler and fully annotation-free.

\paragraph{Interpretation.} The failure is one of \emph{geometric consistency}, not detection: LocA collapses from 51.6 to 9.2 because fused monocular depth does not agree across cameras, so predicted boxes rarely overlap ground truth. Figure~\ref{fig:depth} makes the mechanism visual: the estimated-depth (D4RT) cloud is fragmented and non-planar (top) beside the flat, coherent ground-truth cloud (middle), while the floor-coherence bars (bottom) quantify the collapse---ground-truth points concentrate at the floor with none below it, whereas D4RT scatters an equal share above and below and Metric3D reconstructs no floor at all. Table~\ref{tab:compare} contrasts the two routes. For RGB-only multi-camera 3D tracking in this regime, an explicit geometric 2D$\to$3D lift is far more reliable than learned monocular depth, which---unlike the provided depth used by earlier winners---lacks the cross-view metric consistency that point-cloud detectors require. This reframes the common recommendation to ``move to 3D'': the gap is not the 3D detector but the metric consistency of the geometry fed to it.

\begin{table}
\centering
\caption{Geometry-first lift vs.\ estimated-depth pseudo-LiDAR (test-set 3D HOTA components).}
\label{tab:compare}
\begin{tabular}{lrrrr}
\toprule
Approach & HOTA & DetA & AssA & LocA \\
\midrule
Geometry-first lift (ours) & 13.04 & 10.79 & 16.71 & 51.58 \\
Estimated-depth pseudo-LiDAR & 0.12 & 0.05 & 0.26 & 9.23 \\
\bottomrule
\end{tabular}
\end{table}

\section{Conclusion}

We presented and compared two RGB-only routes to multi-camera 3D tracking for AI City Challenge 2026 Track 1. A geometry-first pipeline---YOLO11x detection, homography lifting, class priors, multi-camera fusion, world-coordinate tracking, and offline tracklet stitching---reaches 13.04 HOTA, while estimated-depth pseudo-LiDAR, the route suggested by prior provided-depth winners, collapses to 0.12 HOTA because monocular depth lacks cross-view metric consistency. Within the geometry pipeline, offline stitching is the only lever that helps, whereas sliced (SAHI) detection, Re-ID, learned MLP lifting, detector ensembling, TTA, and heavy domain randomization do not. The bottlenecks are complementary: detection quality under Sim2Real bounds the geometry pipeline, and localization consistency bounds pseudo-LiDAR. We conclude that, absent inference-time depth, explicit geometry is the more reliable foundation, and that closing the Sim2Real detection-quality gap---not learned monocular depth---is the highest-value next step.

\clearpage
\bibliographystyle{splncs04}
\bibliography{references}

\end{document}